\documentclass[letterpaper]{article}
\usepackage[preprint]{aaai2027}

\usepackage[hyphens]{url}  % DO NOT CHANGE THIS
\usepackage{graphicx} % DO NOT CHANGE THIS
\usepackage{natbib}  % DO NOT CHANGE THIS AND DO NOT ADD ANY OPTIONS TO IT
\usepackage{caption} % DO NOT CHANGE THIS AND DO NOT ADD ANY OPTIONS TO IT
\usepackage{algorithm}
\usepackage{algorithmic}

\usepackage{newfloat}
\usepackage{listings}
\DeclareCaptionStyle{ruled}{labelfont=normalfont,labelsep=colon,strut=off} % DO NOT CHANGE THIS
\floatstyle{ruled}
\newfloat{listing}{tb}{lst}{}
\floatname{listing}{Listing}

\usepackage{booktabs}

\usepackage{amsmath}
\usepackage{booktabs}
\usepackage{tabularx}
\usepackage{array}
\title{T3HG-Editor: Text-driven 3D Human Garment Editing with Body Priors Embedded in SMPL-X}
\author{
	Shaoru Sun\textsuperscript{\rm 1},
	Xingtao Wang\textsuperscript{\rm 1,\rm 2}\corresponding,
	Zihan Ma\textsuperscript{\rm 1},
	Wenrui Li\textsuperscript{\rm 1},\\
	Jiantao Zhou\textsuperscript{\rm 3},
	Debin Zhao\textsuperscript{\rm 1},
	Xiaopeng Fan\textsuperscript{\rm 1,\rm 2,\rm 4}
}

\affiliations{
	\textsuperscript{\rm 1}Faculty of Computing, Harbin Institute of Technology,
	Harbin 150001, China\\
	\textsuperscript{\rm 2}Harbin Institute of Technology Suzhou Research Institute,
	Suzhou 215000, China\\
	\textsuperscript{\rm 3}Department of Computer and Information Science,
	University of Macau, Macau, China\\
	\textsuperscript{\rm 4}Peng Cheng Laboratory, Shenzhen 518000, China\\
	24b903087@stu.hit.edu.cn,
	xtwang@hit.edu.cn,
	120l020523@stu.hit.edu.cn,
	liwr@stu.hit.edu.cn\\
	jtzhou@um.edu.mo,
	dbzhao@hit.edu.cn,
	fxp@hit.edu.cn
}

\begin{document}

\maketitle

\begin{abstract}
While 3D Gaussian Editing (3DGE) has seen substantial progress, text-driven 3D human garment editing remains largely underexplored. Existing 3DGE works typically follow a paradigm that applies 2D editing techniques to multi-view rendered images and updates 3D Gaussians based on the modified images. Extending such methods to 3D human garment editing suffers from low-fidelity outcomes, caused by introduced distortions and garment inconsistencies. A promising breakthrough opportunity arises from the SMPL eXpressive (SMPL-X) model that embodies rich prior information for virtual humans. Motivated by this insight, we propose a text-driven 3D human garment editor termed T3HG-Editor, which delivers high-fidelity and garment consistent results by leveraging geometry and joint priors embedded in SMPL-X. Specifically, T3HG-Editor contains three stages, namely obtainment of editable Gaussians, garment consistent editing, and Gaussian updating with overflow pruning. The obtainment of editable Gaussians begins with seeding Gaussians along SMPL-X normals to generate sufficient near surface Gaussians, followed by a 2D mask constraint that precisely localizes the target Gaussians to be edited. The garment consistent editing aggregates tokens corresponding to the same SMPL-X vertex across multiple views and propagates them to their original views, enforcing garment consistency without requiring additional training. Gaussian updating with overflow pruning employs a Signed Distance Function (SDF) defined on SMPL-X to construct a human distance field, which is then integrated with a 2D semantic mask to prune overflowing Gaussians, thus preventing contamination of non-target regions. Experiments on multiple subjects and diverse garment types demonstrate that T3HG-Editor outperforms state-of-the-art methods in both editing quality and garment consistency.
\end{abstract}

% Uncomment the following to link to your code, datasets, an extended version or similar.
% You must keep this block between (not within) the abstract and the main body of the paper.
% Make sure that you do not de-anonymize yourself with these links.
% \begin{links}
%    \link{Code}{https://ssr98-rgb.github.io/T3HG-Editor/}{https://ssr98-rgb.github.io/T3HG-Editor/}
%     \link{Datasets}{https://ssr98-rgb.github.io/T3HG-Editor/}{https://ssr98-rgb.github.io/T3HG-Editor/}
%     \link{Extended version}{https://ssr98-rgb.github.io/T3HG-Editor/}{https://ssr98-rgb.github.io/T3HG-Editor/}
% \end{links}

\begin{figure}[!t]
	\centering
	\includegraphics[width=\linewidth]{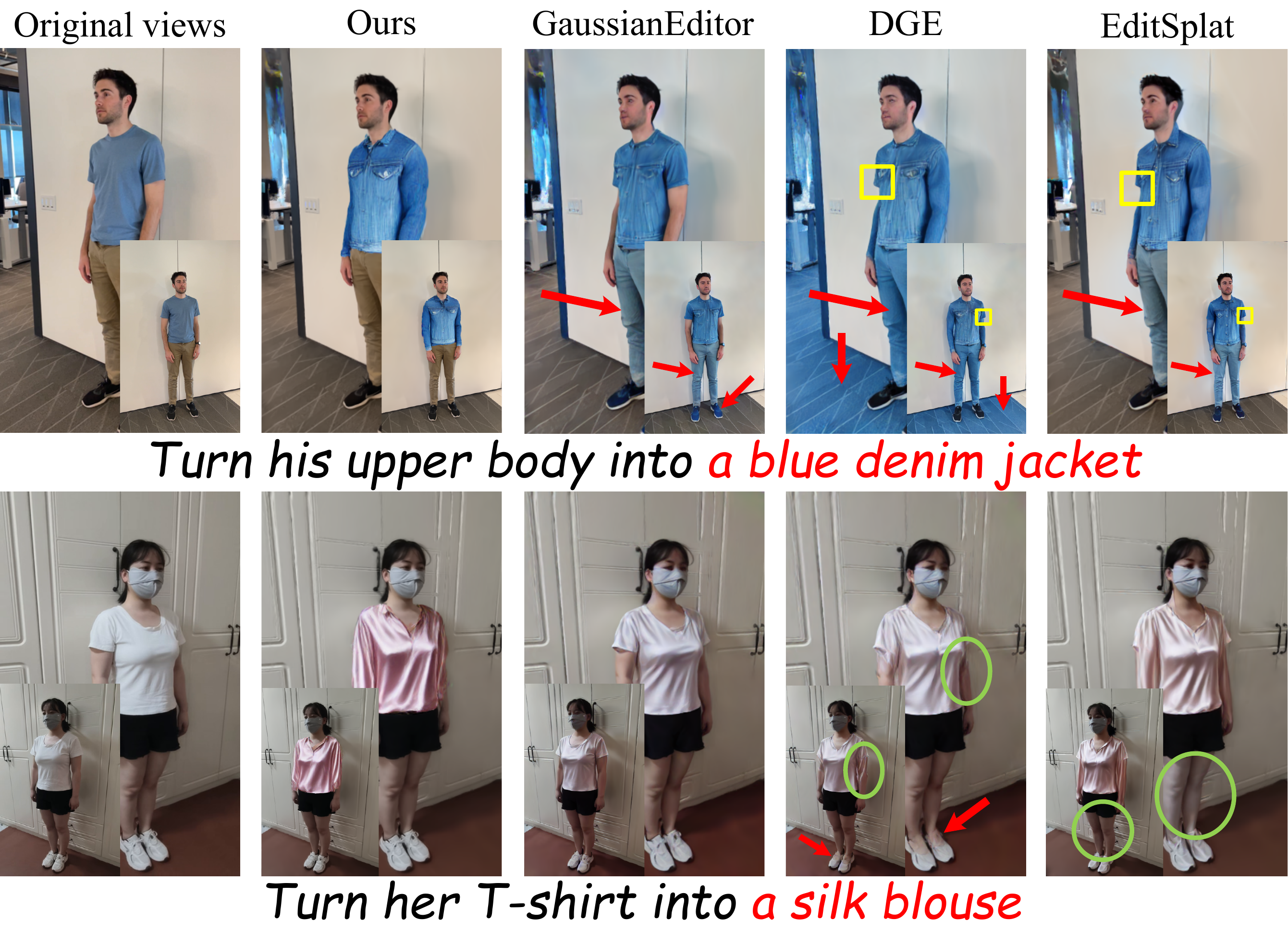}
	\caption{In the top row, existing methods alter the background and pants colors (\textcolor{red}{red arrows}) while retaining short-sleeve details on the denim jacket (\textcolor{yellow}{yellow box}). In the bottom row, garment inconsistencies appear around the sleeves and shoes (\textcolor{green}{green ellipses}).}
	\label{wuran}
\end{figure}

\section{Introduction}

3D human garment editing holds significant value in meeting personalized demands across various domains like AIGC asset reuse, film production, e-commerce display, and more \cite{11021338,11024125,ZHANG201640}. Recent advances in 2D diffusion models \cite{10740586,10934729,ho2020denoising} and 3D Gaussian splatting (3DGS) \cite{kerbl20233d} have greatly advanced the development of 3D Gaussian Editing (3DGE). However, 3D human garment editing, which requires fine-grained and garment consistent modifications without disrupting background or human structure, remains underexplored due to the diversity of garment types and the complexity of their geometries \cite{sun2025deep,chen2024gaussianvton}. Existing 3DGE works typically apply 2D editing techniques to multi-view rendered images and subsequently update 3D Gaussians based on the modified images \cite{vachha6instruct,karim2024free}. A straightforward extension of such methods to human garments suffers from low-fidelity outcomes, primarily on account of two key issues. As shown in Figure \ref{wuran}, on the one hand, complex non-rigid deformations in garment regions may lead to inaccurate shape adaptation, while unintended Gaussian contamination may occur in non-garment regions. On the other hand, garment diversity and independent editing across different views may result in inconsistencies in garment shape, texture, and appearance.

The SMPL-X (Skinned Multi-Person Linear Model with eXpressive hands and face) model \cite{pavlakos2019expressive,loper2023smpl}, which is widely adopted in the virtual human domain, encapsulates rich prior information including human body geometry, pose joints, and garment-body interaction constraints \cite{lin2025siavatar,svitov2024haha,shao2025degas}. The priors embedded in SMPL-X enable restricting edits within target garment regions and establishing consistency constraints across views. Motivated by this insight, we propose a text-driven 3D human garment editor termed T3HG-Editor. T3HG-Editor leverages SMPL-X's body geometry prior to constrain Gaussian position adjustment and target-non-target region decoupling, and utilizes the pose joint prior to achieve cross-view consistent feature fusion. T3HG-Editor contains three stages, namely obtainment of editable Gaussians, garment consistent editing, and Gaussian updating with overflow pruning. The obtainment of editable Gaussians begins by seeding 3D Gaussians along the normals of the SMPL-X model’s surface, thus anchoring the Gaussians to the geometric structure of the human body. Then, 2D mask-based filtering eliminates Gaussians outside the target garment region to obtain accurately positioned Gaussians. For garment consistent editing, T3HG-Editor aggregates feature tokens associated with the same SMPL-X vertex across views. By tying token aggregation to the unified geometric framework of SMPL-X for information sharing, garment consistent editing aligns Gaussian representations across different views without requiring additional model training. Gaussian updating with overflow pruning follows a dual-constraint paradigm. One constraint is a 3D human distance field, primarily derived from a Signed Distance Function (SDF) defined on the SMPL-X model, while the other is 2D semantic masks. This dual-constraint paradigm confines Gaussians to the intended regions, thus preventing the contamination of unintended regions. In summary, the contributions of this work are as follows:

\begin{itemize}
	\item We propose a text-driven 3D human garment editor (T3HG-Editor) based on the SMPL-X model. To the best of our knowledge, T3HG-Editor is an early systematic exploration of text-driven 3DGS editing specifically designed for human garment editing.
	
	\item Based on SMPL-X’s body geometric prior, we propose Gaussian seeding along the surface normals and target-non-target region decoupling, solving shape deviations in the target garment region and unintended Gaussian contamination in non-target regions.
	
	\item Based on the pose joint prior, we design a garment consistent editing scheme by aggregating tokens corresponding to the same SMPL-X vertex across key views.
	
	\item We validate T3HG-Editor on garment editing tasks across different genders, showing that it better preserves fidelity in both target garment and non-target regions while simultaneously achieving superior garment consistency over state-of-the-art editing methods.
\end{itemize}

\section{Related work}
\label{gen_inst}

\textbf{3D Human Gaussian Splatting with SMPL-X.} By providing a parametric skeleton, vertices, and meshes, SMPL-X has been widely adopted in 3D human Gaussian modeling \cite{prospero2025gst,hu2024gauhuman}. HumanGaussian \cite{liu2024humangaussian} initializes Gaussian positions with SMPL-X meshes and employs a dual-branch SDS scheme for Gaussian updating. DreamWaltz-G \cite{huang2025dreamwaltz} uses the SMPL-X skeleton as a conditioning signal and SMPL-based Gaussian initialization to reduce multi-face artifacts and implausible poses. ToMiE \cite{zhan2024tomie} identifies garment regions not covered by the base model and grows external data-driven joints for decoupled animation of loose garments and hand-held objects. GaussianAvatar \cite{hu2024gaussianavatar} binds animatable Gaussians to the SMPL-X skeleton, applies forward LBS for reposing, and jointly refines SMPL-X pose parameters for monocular human reconstruction. Across these works, SMPL-X plays a central role in 3D human generation \cite{pan2024humansplat}, reconstruction \cite{li2024gaussianbody}, and animation \cite{kocabas2024hugs} with Gaussian representations. In this paper, we adopt SMPL-X priors to facilitate garment editing of 3D human Gaussians.

\begin{figure*}[!t]
	\centering
	\includegraphics[width=\linewidth]{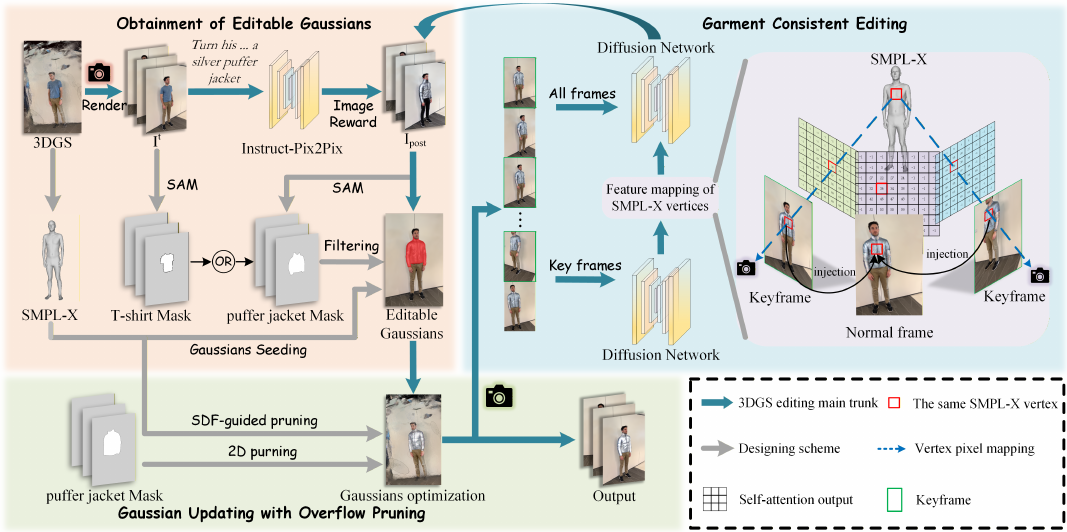}
	\caption{The overall framework of \textbf{T3HG-Editor}, which consists of three components: (1) Obtainment of Editable Gaussians; (2) Garment Consistent Editing; and (3) Gaussian Updating with Overflow Pruning.}
	\label{framework}
\end{figure*}

\textbf{Text-driven 3D Editing.} The scarcity of large-scale 3D training data has motivated most text-driven 3D editing methods to leverage pretrained 2D diffusion models \cite{couairon2022diffedit,li2024scenedreamer360,narasimhaswamy2024handiffuser,chen2024consistdreamer}. InstructPix2Pix (IP2P) \cite{brooks2023instructpix2pix}, conditioned on an input image and text instruction, is a widely adopted editing backbone. Instruct-NeRF2NeRF \cite{haque2023instruct} iteratively edits rendered views with IP2P and uses them to update a NeRF representation. However, the implicit representation of NeRF makes local editing difficult to control and computationally expensive. With the emergence of 3D Gaussian Splatting, GaussianEditor \cite{chen2024gaussianeditor} back-projects 2D editing masks to label target Gaussians and employs anchor constraints to regulate Gaussian densification. Nevertheless, independently editing individual views often causes cross-view inconsistencies. Subsequent methods therefore introduce geometric or feature-level constraints. View-consistent 3D editing \cite{wang2024view} exploits cross-view geometric correspondence, while DGE \cite{chen2024dge} jointly attends to key views and propagates edited features to other views via epipolar constraints. EditSplat \cite{lee2025editsplat} incorporates multi-view depth information and attention-guided Gaussian pruning, whereas GaussCtrl \cite{wu2024gaussctrl} jointly edits multiple views with depth conditioning and attention alignment. Other approaches align instruction-related regions with Gaussian representations to support localized updates \cite{wang2024gaussianeditor}.

\section{Method}
\label{headings}

As illustrated in Figure \ref{framework}, T3HG-Editor comprises three stages: obtainment of editable Gaussians, garment consistent editing, and Gaussian updating with overflow pruning. The first stage combines SMPL-X-based Gaussian seeding with mask-based filtering to obtain editable Gaussians. The second stage establishes cross-view correspondence through SMPL-X-vertex guided attention and feature propagation. The third stage updates the editable Gaussians while applying 2D and 3D pruning to prevent contamination of unintended regions. The following subsections describe these stages.

\subsection{Obtainment of Editable Gaussians}
\label{CVG}

Existing 3D editing methods obtain editable Gaussians under the guidance of masks or attention weights. However, such schemes are unsuitable for garment editing because garment coverage may change after editing. Restricting Gaussian updating to the pre-edit region overlooks newly extended parts, whereas focusing on the post-edit region ignores contracted parts. To address these limitations, T3HG-Editor seeds Gaussians along the SMPL-X normals to generate sufficient Gaussians proximate to the body surface, followed by 2D mask filtering to precisely position the target Gaussians.

\begin{figure}[!t]
	\centering
	\includegraphics[width=\linewidth]{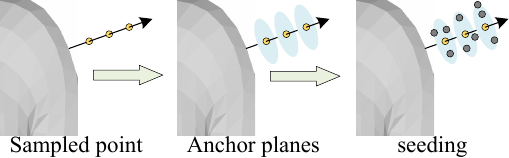}
	\caption{SMPL-X-based Gaussian seeding. The yellow circles indicate the points sampled along the normals, while the gray circles are the positions of the seeded Gaussians.}
	\label{seeding}
\end{figure}

\textbf{SMPL-X-based Gaussian Seeding.} This operation can generate a dense set of editable Gaussians based on an SMPL-X mesh, effectively mitigating shape deviations that arise after editing operations. As shown in Figure \ref{seeding}, given an SMPL-X mesh, the SMPL-X-based Gaussian Seeding operation begins by uniformly sampling $J$ points within a distance $D_{\max}$ along the normal direction of each vertex. Then, each sampled point is used to anchor a plane whose normal is aligned to the corresponding vertex normal. Afterwards, for each plane, a set of seeded points is generated through random sampling centered at the corresponding anchor point and confined within radius $r$. Formally, the position of the seeded points is expressed as follows: 
\begin{equation}
	\mathbf{x}_{i,j} = \mathbf{v}_i + s_j \cdot \mathbf{n}_i + u \cdot \mathbf{t}_{i,j},
	\quad u \sim \mathcal{U}[0,r],
\end{equation}
\begin{equation}
	s_j = \frac{j-1}{J-1} \cdot D_{\max}, \quad j=1,2,\ldots,J.
\end{equation}
Here, $\mathbf{n}_i$ represents the normal of vertex $\mathbf{v}_i$, $\mathbf{t}_{i,j}$ denotes a random unit vector, $u$ denotes a random number sampled from the interval $[0,r]$. Finally, each seeded point is assigned appearance and geometric properties through KNN \cite{cover1967nearest} attribute aggregation, thus defining the initial state of the seeded Gaussians. The color parameter $\mathbf{c}_{i,j}$, scaling factor $\mathbf{s}_{i,j}$ and opacity $\alpha_{i,j}$ are initialized as the average of the corresponding attributes from the several nearest Gaussians, while the rotation parameter $\mathbf{q}_{i,j}$ inherits directly from the single nearest Gaussian in the original model.

\begin{figure}[!t]
	\centering
	\includegraphics[width=\linewidth]{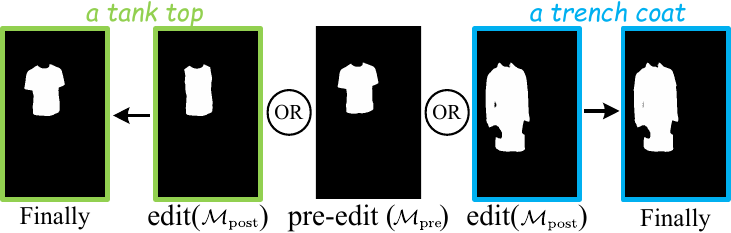}
	\caption{Obtaining target masks for garments that shrink or expand after editing.}
	\label{loc}
\end{figure}

\textbf{Mask-based filtering.} For the seeded Gaussians, pre-edit and post-edit garment changes are jointly considered to accurately localize the editable Gaussians. Specifically, the 3DGS scene is rendered from $T$ views to obtain $\{I_{\text{pre}}^{t}\}_{t=1}^{T}$, and Segment Anything Model (SAM) \cite{kirillov2023segment} is applied to derive the pre-edit garment mask $\mathcal{M}_{\text{pre}} $. Then, $\{I_{\text{pre}}^{t}\}_{t=1}^{T}$ together with the text prompt are fed into InstructPix2Pix (IP2P) \cite{brooks2023instructpix2pix} to generate the edited images. IP2P generates diverse results across viewpoints, and such diversity facilitates selecting the outputs that align with the textual description, thereby establishing the garment’s overall shape. Accordingly, ImageReward \cite{xu2023imagereward} is used to evaluate the editing results, and the top $N$ images are selected as supervision for subsequent Gaussian updates, denoted by $\{I_{\text{post}}^{n}\}_{n=1}^{N}$. Subsequently, these images are fed into SAM to obtain the post-edit garment mask $\mathcal{M}_{\text{post}}$. Finally, the filtering mask is obtained by
$\mathcal{M}_{\text{filter}}=\mathcal{M}_{\text{pre}}\lor\mathcal{M}_{\text{post}}$, which accommodates garment expansion or shrinkage, as illustrated in Figure \ref{loc}. The filtered Gaussians are assigned semantic masks that are retained during subsequent splitting or duplication.

According to the back-projection formulation in GaussianEditor \cite{chen2024gaussianeditor}, the computed Gaussian weights determine whether a Gaussian is assigned a semantic mask, thereby integrating garment shapes from all edited views. To obtain a unified 2D garment representation, the labeled Gaussians are separately rendered to the selected camera views, and the resulting 2D mask $\mathcal{M}_{\text{final}}$ is used both as the loss region and as the supervision region for subsequent pruning. The editing loss is defined as follows:
\begin{equation}
	\mathcal{L}_{\text{edit}} = \min \sum_{n=1}^{N} 
	\left\| I^{n}_{\text{post}}(\mathcal{M}_{\text{final}}) - I^{\prime}_{n}(\mathcal{M}_\text{final}) \right\|.
\end{equation}
Here, $I_n^{\prime}$ denotes the rendered image of the current 3D Gaussians at the $n$-th view. Through the above mechanism, T3HG-Editor can precisely obtain the editable Gaussians, thereby establishing a reliable foundation for subsequent multi-view garment consistency enhancement and pruning.

\subsection{Garment Consistent Editing}
\label{SGC}
The independent single-view editing inevitably introduces garment inconsistencies in texture and fine details. Therefore, a garment consistency enhancement mechanism is designed comprising two core modules, SMPL-X-vertex guided attention and SMPL-X-vertex guided feature propagation, achieving unified semantic and visual coherence across views.

In the context of human garment editing, the fitted SMPL-X mesh carries effective structural priors for establishing cross-view correspondences. Given the camera parameters, SMPL-X vertices are rendered into each view and every pixel patch is annotated with its associated vertex index. This yields a bidirectional lookup between pixel patches and mesh vertices, furnishing a unified and queryable coordinate reference for subsequent cross-view feature alignment. 

\textbf{SMPL-X-vertex Guided Attention.} Unlike editing a single image, T3HG-Editor jointly edits multi-view images by concatenating the tokens corresponding to the same SMPL-X vertex before applying attention, thereby enhancing consistency. Specifically, to reduce computational cost, one key view out of every five views is selected in order and all key views are fed into the U-Net simultaneously. Then, within the U-Net’s self-attention module, for each key view $m$ with queries $\{Q_m\}_{m=1}^M$, keys $\{K_m\}_{m=1}^M$, and values $\{V_m\}_{m=1}^M$, the tokens corresponding to the same vertex $l$ across key views are determined from the vertex–pixel index list. Next, these tokens for the same vertex are concatenated and processed through self-attention, enabling each view to attend to the features of other views corresponding to the same SMPL-X vertex. The formulation is as follows:

\begin{equation}
	\resizebox{0.96\columnwidth}{!}{$\displaystyle
		\mathrm{STAttn}\!\bigl(Q^{l},K^{l},m\bigr)
		=
		\operatorname{Softmax}\!\left(
		\frac{Q_m^{l}\cdot[K^{l}_1,\ldots,K^{l}_M]}{\sqrt{d}}
		\right)
		$},
\end{equation}
\begin{equation}
	\Phi^{l}_m
	= \operatorname{STAttn}\!\bigl(Q^{l}, K^{l}, m\bigr)
	\cdot [V^{l}_1, \ldots, V^{l}_M].
	\label{feature}
\end{equation}
Here, $d$ denotes the dimension used for computing queries and keys. $l \in \{1, \cdots, L\}$ denotes the same vertex across key views. After computing cross-view attention for each vertex, the output features are returned to their positions. Finally, tokens that cannot be matched to the same vertex across views, along with background tokens, are concatenated and processed through self-attention across key views. 

By performing the vertex-level cross-view attention only within the selected key views and processing vertex groups in batches, the memory footprint and computational overhead can be effectively controlled. This operation explicitly enhances the semantic and textural consistency of the same vertex across multiple key views, providing a well-aligned foundation for subsequent feature propagation.

\textbf{SMPL-X-vertex Guided Feature Propagation.} For the remaining views, the features from the key views are propagated according to the principle of sharing the same vertex, thereby ensuring consistency across views. The vertex-pixel index list is used to query and compute the tokens corresponding to the same vertex in both normal views and key views. Then, the features $\Phi^{l}_m$ obtained from Eq. (\ref{feature}) are fused through weighted aggregation and injected into the remaining views as self-attention outputs, as formulated below:

\begin{equation}
	\psi^{l}_n = w_1 \cdot \Phi^{l}_1 + \cdots + w_M \cdot \Phi^{l}_M.
\end{equation}
The weight $w_m$ of each key view is determined by normalizing the number of vertices. It shares with the $n$-th normal view, where more shared vertices imply closer views and thus higher weights. For the remaining tokens without corresponding vertices, the cosine similarity is computed between the tokens of the normal and key views, and the most similar tokens are selected as learning targets.

Through the above mechanism, SMPL-X establishes information association and mutual learning across multiple views, thereby improving consistency. In addition, some vertices may be occluded or invisible in certain views but remain visible in others. When performing cross-view self-attention and feature learning, to avoid losing information from invisible vertices on the human body, the tokens of the same vertex are processed together with their neighboring features for attention and feature learning.

\subsection{Gaussian Updating with Overflow Pruning}
\label{DGP}
During Gaussian updating, some Gaussians may deviate from the target regions, causing contamination of non-target regions. To address this issue, 2D and 3D pruning are designed by combining a Signed Distance Function (SDF)-based human distance field with 2D masks, thereby pruning Gaussians that overflow beyond the target editing regions.

\textbf{2D Pruning.} After periodic updating, the Gaussians labeled with the semantic mask are rendered from the selected camera views, and the resulting current Gaussian mask map is compared with $\mathcal{M}_{\text{final}}$ to identify out-of-bound pixel areas. These pixel areas are then back-projected into 3D space to track the Gaussians that have moved beyond the target region for pruning. It should be noted that this process applies only to the labeled Gaussians.

\begin{figure}[!t]
	\centering
	\includegraphics[width=1\linewidth]{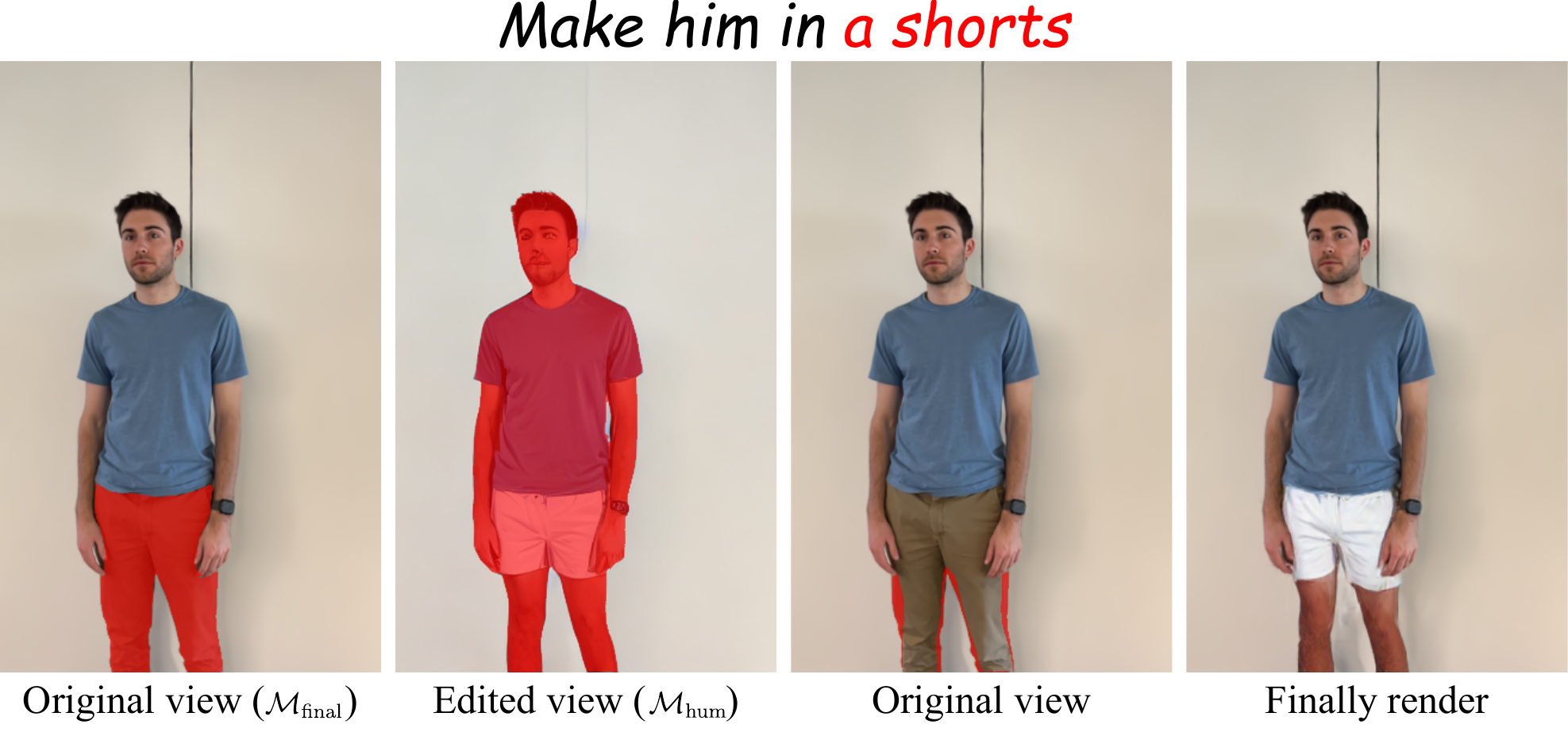}
	\caption{The original pants and edited human are segmented to identify pruning regions, followed by Gaussian pruning and updating to obtain the final rendering.}
	\label{shrink}
\end{figure}

For garment shrinkage, such as ``\textit{Make him in a shorts}'', Gaussians corresponding to the lower pants gradually become transparent or are pruned. However, gradient-based updating alone may introduce artifacts. As shown in Figure \ref{shrink}, a human segmentation mask $\mathcal{M}_{\text{hum}}$ is generated for the edited human and excludes the lower pants, whereas $\mathcal{M}_{\text{final}}$, obtained by the logical OR of the pants and shorts masks, retains the original pants shape. Comparing $\mathcal{M}_{\text{hum}}$ with $\mathcal{M}_{\text{final}}$ efficiently localizes and prunes the Gaussians associated with the missing lower pants.

\textbf{3D Pruning.} Due to viewpoint limitations and occlusion, some Gaussians may drift away from the human body while their 2D projections remain within $\mathcal{M}_{\text{final}}$. To address this, an SMPL-X-derived SDF constraint is imposed in 3D space. For Gaussians generated through splitting or duplication, the signed distance from each centroid to the SMPL-X surface is computed, and those exceeding a predefined threshold $\beta$ are pruned. This constraint removes out-of-body drift that is difficult to detect in 2D projections.

Dual-domain pruning can significantly reduce the risk of noisy Gaussians contaminating non-target regions. However, given the large number of Gaussians, missed drifting Gaussians may still cause local contamination that propagates “from points to surfaces”. To prevent such cases, a loss is introduced in non-target regions to keep their appearance consistent with the original scene, as formulated below:

\begin{equation}
	\mathcal{L} = \mathcal{L}_{\text{edit}} +
	\min \sum_{n=1}^{N} 
	\left\| I^{n}_{\text{pre}}(\overline{\mathcal{M}_{\text{final}}}) - I_n^{\prime}(\overline{\mathcal{M}_{\text{final}}}) \right\|.
\end{equation}
Here, $\overline{\mathcal{M}_{{\text{final}}}}$ denotes the complement of $\mathcal{M}_{{\text{final}}}$, representing the non-target regions. Gradients are computed between the rendered images $I_n^{\prime}$ of the current 3DGS and the original $I^{n}_{\text{pre}}$ in the unintended regions, thereby preserving these regions unchanged. This is also facilitated by the accurate localization of the 2D mask.

\begin{figure*}[!t]
	\centering
	\includegraphics[width=\linewidth]{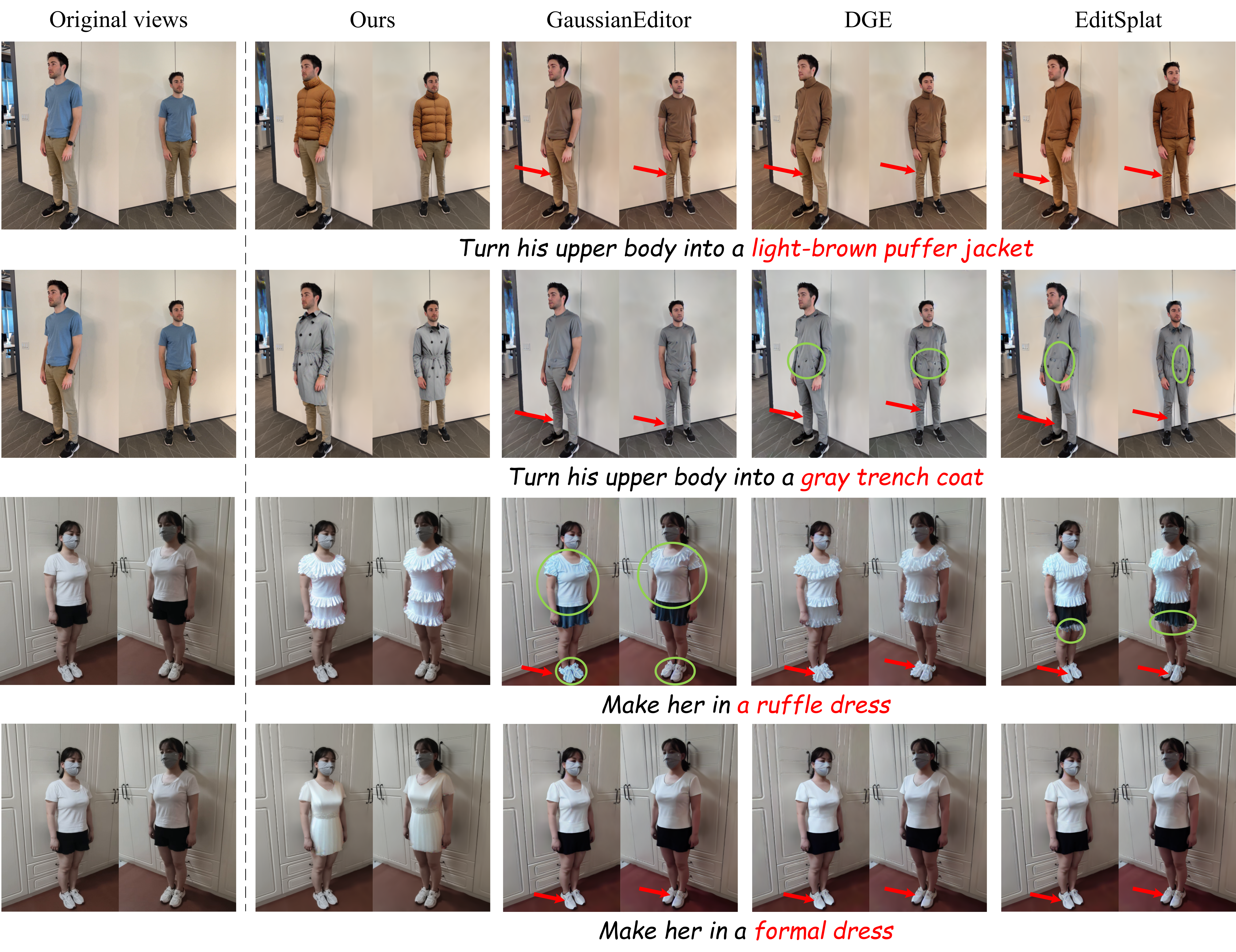}
	\caption{Qualitative comparison of garment editing. The regions that require attention have been annotated in the figure. The \textcolor{red}{red} arrows indicate unintended region changes, while the \textcolor{green}{green} ellipses highlight garment inconsistencies.}
	\label{man1}
\end{figure*}

\section{Experiments}
\label{Experiments}

\subsection{Experimental Setup}
\textbf{Implementation Details.} T3HG-Editor adopts 3D Gaussian Splatting (3DGS) to represent humans and employs InstructPix2Pix \cite{brooks2023instructpix2pix} as the image editing backbone. For each scene, the number of selected views $N$ is set to 20, and the Gaussian updating is performed for 1200 iterations. Following the recommendations of IN2N \cite{haque2023instruct}, both LPIPS and L1 losses are utilized to update the 3DGS human scenes. The maximum seeding distance $D_{\max}$ and the SDF-based distance threshold $\beta$ are determined according to the size and thickness of the target garment to be edited, and are typically set within the range of 0 to $0.1\,\mathrm{m}$. All experiments are conducted on a workstation equipped with an NVIDIA GeForce RTX 4080 SUPER GPU (32 GB VRAM), and the editing process takes approximately 7 to 9 minutes.

\textbf{Compared Methods}. T3HG-Editor is compared with state-of-the-art 3D scene editing approaches, including GaussianEditor \cite{chen2024gaussianeditor}, DGE \cite{chen2024dge}, and EditSplat \cite{lee2025editsplat}. All methods adopt InstructPix2Pix as the image editor and have been applied to garment edits in human scenes. Among them, GaussianEditor is a representative approach in this field. DGE and EditSplat further introduce consistency enhancing strategies, making them competitive methods for comparison.

\textbf{Evaluation Metrics.} The quantitative evaluation follows the criteria introduced in \cite{he2024customize,wang2024view}. Specifically, 20 camera poses are randomly rendered from the edited 3DGS, and their similarity to the target text is computed using CLIP \cite{radford2021learning}. At the same time, the CLIP directional similarity is computed to evaluate the alignment between the image editing direction and the textual editing direction. All methods are assessed across ten text prompts. In addition, one garment editing case is selected to compare the trends of CLIP and CLIP directional similarity, thereby illustrating the efficiency of updating and the trajectory of performance improvement.

\subsection{Qualitative Comparisons}
\label{Qualitative Comparisons}
Figure \ref{man1} presents qualitative results of garment editing on male and female subjects. As we can see, unintended changes occur in the regions indicated by the red arrows for GaussianEditor, DGE, and EditSplat. In addition, for the text prompts ``\textit{Turn his upper body into a gray trench coat}'', DGE and EditSplat exhibit garment inconsistencies in sleeve or button editing, as indicated by the green ellipses. In contrast, T3HG-Editor confines modifications strictly to the target garment regions by jointly considering masks before and after editing, thereby avoiding contamination of irrelevant areas. Moreover, when editing a T-shirt into a long-sleeved coat, the three baseline methods still leave visible short-sleeve contours. For large garments, such as the prompt ``\textit{Turn his upper body into a light-brown puffer jacket}'', they do not capture the bulky texture characteristic of puffer jackets. These limitations stem from the insufficient distribution of Gaussians in garment-related regions such as sleeves, hems, and other structural components in the original 3D scenes. For female subjects, similar limitations remain in the compared methods, and they fail to produce edits consistent with textual instructions and often alter Gaussian features in unintended regions, as indicated by the red arrows. Under the prompt ``\textit{Turn her T-shirt into a silk blouse}'', as indicated by the green ellipses, DGE yields inconsistent sleeves between the left and right arms, while EditSplat produces mismatched colors on the legs. With the prompt ``\textit{Make her in a formal dress}'', all three baselines fail to generate results resembling a formal dress, producing outputs largely similar to the original garment. 

By contrast, T3HG-Editor leverages SMPL-X normals to seed Gaussians, effectively supplementing missing regions and precisely obtaining Gaussians in the target region using 2D masks, thereby validating the effectiveness of the proposed obtainment of editable Gaussians scheme. Moreover, T3HG-Editor achieves garment consistent editing through cross-view attention among tokens corresponding to the same SMPL-X vertices, while pruning overflowing Gaussians to prevent contamination of unintended regions.

\subsection{Quantitative Comparisons}
Table \ref{tab:clip_results} presents the CLIP similarity, CLIP directional similarity, and average running time of T3HG-Editor and the compared approaches across two scenarios and ten text prompts. The results show that T3HG-Editor outperforms state-of-the-art approaches in both CLIP similarity and CLIP directional similarity. T3HG-Editor requires time for Gaussian seeding and cross-view attention computation among tokens corresponding to the same SMPL-X vertices within the U-Net. Therefore, it takes more time than DGE, while remaining competitive with the other methods.

Table \ref{RSNP} reports the average PSNR, SSIM, and LPIPS over ten text instructions. For each method, 16 of the 20 edited views are used for 3DGS reconstruction, while the remaining four are used for testing. The rendered test views are compared with the edited images in the target garment regions and with the original images in the non-target regions. T3HG-Editor achieves the best SSIM and LPIPS in the target garment regions, demonstrating improved multi-view garment consistency. Its PSNR is slightly lower than that of DGE because DGE fails on some instructions and produces less detailed textures. In the non-target garment regions, T3HG-Editor performs best on all metrics, benefiting from accurate obtainment of editable Gaussians and overflow pruning.

\begin{table}[t]
	\centering
	\footnotesize
	\setlength{\tabcolsep}{1pt}
	\renewcommand{\arraystretch}{1.15}
	
	\begin{tabular*}{\columnwidth}{
			@{\extracolsep{\fill}}lccc@{}
		}
		\toprule
		Method
		& $\mathrm{CLIP}_{\mathrm{sim}}$
		& $\mathrm{CLIP}_{\mathrm{dir}}$
		& Time (min)
		\\
		\midrule
		
		GaussianEditor (CVPR 2024)
		& 0.2524
		& 0.0889
		& $\sim$8
		\\
		
		DGE (ECCV 2024)
		& 0.2573
		& 0.1445
		& \textbf{$\sim$4}
		\\
		
		EditSplat (CVPR 2025)
		& 0.2554
		& 0.1674
		& $\sim$24
		\\
		
		T3HG-Editor (Ours)
		& \textbf{0.2660}
		& \textbf{0.2394}
		& $\sim$8
		\\
		
		\bottomrule
	\end{tabular*}
	
	\caption{Quantitative comparisons. The best results are highlighted in \textbf{bold}.}
	\label{tab:clip_results}
\end{table}

\begin{table}[!t]
	\centering
	\footnotesize
	\setlength{\tabcolsep}{4pt}
	\renewcommand{\arraystretch}{1.30}
	
	\begin{tabular*}{\columnwidth}{@{\extracolsep{\fill}}lccc@{}}
		\toprule
		Method & PSNR$\uparrow$ & SSIM$\uparrow$ & LPIPS$\downarrow$ \\
		\midrule
		
		\textit{Target Garment Region} & & & \\[1pt]
		
		GaussianEditor (CVPR 2024) 
		& 19.4225
		& 0.5290
		& 0.2904 \\
		
		DGE (ECCV 2024) 
		& \textbf{22.0834}
		& 0.5783
		& 0.2085 \\
		
		EditSplat (CVPR 2025) 
		& 20.9711
		& 0.5731
		& 0.2157 \\
		
		T3HG-Editor (Ours)
		& 21.2198
		& \textbf{0.5796}
		& \textbf{0.1738} \\
		
		\midrule
		
		\textit{Non-target Garment Region} & & & \\[1pt]
		
		GaussianEditor 
		& 20.0683
		& 0.7963
		& 0.2279 \\
		
		DGE 
		& 23.3919
		& 0.8701
		& 0.1644 \\
		
		EditSplat 
		& 19.7180
		& 0.7392
		& 0.3143 \\
		
		T3HG-Editor (Ours)
		& \textbf{28.8880}
		& \textbf{0.9300}
		& \textbf{0.0828} \\
		
		\bottomrule
	\end{tabular*}
	\caption{Quantitative comparisons on the target and non-target garment regions.}
	\label{RSNP}
\end{table}

\begin{figure}[!t]
	\centering
	\includegraphics[width=\linewidth]{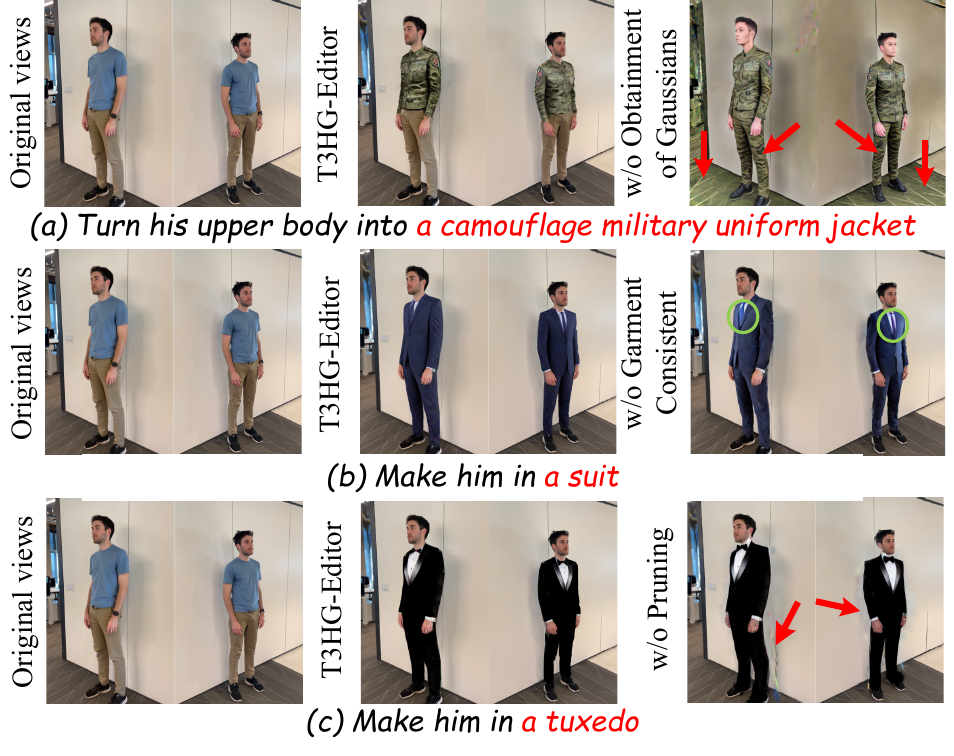}
	\caption{\textbf{Ablation Study.} (a)--(c) illustrate the effects of editable Gaussian obtainment, garment consistency enforcement, and Gaussian overflow pruning, respectively. The \textcolor{red}{red} arrows indicate unintended changes, while the \textcolor{green}{green} ellipses highlight garment inconsistencies.}
	\label{xiaorong}
\end{figure}

\subsection{Ablation Study}
To validate the proposed obtainment of editable Gaussians, garment consistent editing, and pruning of Gaussian overflow, ablation studies are conducted on each mechanism. 

First, as shown in Figure \ref{xiaorong}(a), removing the obtainment of editable Gaussians causes facial distortion and background corruption. The edited garment retains a T-shirt contour because the original scene lacks Gaussians around the sleeves, demonstrating the necessity of Gaussian seeding and the effectiveness of editable Gaussian obtainment. 

Second, Figure \ref{xiaorong}(b) demonstrates the contribution of garment consistent editing. Without this mechanism, the tie exhibits inconsistent colors across views, and artifacts appear on the suit, as highlighted by green ellipses. Applying this strategy produces a uniform tie color and fewer artifacts, confirming its effectiveness in enhancing garment consistency.

Finally, Figure \ref{xiaorong}(c) verifies the effectiveness of Gaussian overflow pruning. Without pruning, numerous noisy Gaussians and artifacts appear around the human body, as indicated by the red arrows, whereas the proposed pruning effectively suppresses these issues.

\section{Conclusion}

To address complex shape transformations and garment inconsistencies in 3DGS-based garment editing, we propose T3HG-Editor, a text-driven 3D human garment editor. It seeds Gaussians along SMPL-X surface normals and filters them using fused pre- and post-edit masks to localize editable regions. Cross-view attention aggregates features associated with the same SMPL-X vertices, while an SDF-based distance field and 2D masks prune overflowing Gaussians. Experiments on garments show that T3HG-Editor outperforms state-of-the-art methods qualitatively and quantitatively.

T3HG-Editor relies on IP2P, whose ability to handle complex text instructions remains limited, while simultaneous multi-view editing increases runtime. Future work will focus on accelerating editing and supporting more diverse garment modifications with complex instructions.

\section*{Acknowledgments}

This work was supported in part by the National Key R\&D Program of China (2025ZD1601300), the National Natural Science Foundation of China (NSFC) under grants 62402138 and 624B2049.

\bibliography{aaai2027}

% Check whether the conference requires a reproducibility checklist to be included in the paper.
% If so, you can uncomment the following line and ajust the path to include it.
% \input{ReproducibilityChecklist.tex}

\end{document}